\documentclass[runningheads]{llncs}
\usepackage[T1]{fontenc}
\usepackage{graphicx}
\usepackage{xcolor}
\usepackage{amsmath}
\usepackage{multirow}
\usepackage{float}
\usepackage{subfigure}
\usepackage{caption}
\usepackage{tikz}
\usetikzlibrary{shapes, arrows.meta}
\begin{document}
\title{An Automated Approach to Collecting and Labeling Time Series Data for Event Detection Using Elastic Node Hardware}
\titlerunning{Automated Time Series Data Collection and Labeling for Event Detection}

\author{Tianheng Ling \and
Islam Mansour \and
Chao Qian \and
Gregor Schiele }
\authorrunning{Ling et al.}

\institute{Intelligent Embedded Systems Lab, University of Duisburg-Essen, \\ 
47057, Duisburg, Germany \\
\email{\{tianheng.ling, chao.qian, gregor.schiele\}@uni-due.de} \\
\email{islam.mansour@stud.uni-due.de}
}
\newcommand{\edit}[1]{\textcolor{red}{#1}}
\maketitle 

\begin{abstract}

Recent advancements in IoT technologies have underscored the importance of using sensor data to understand environmental contexts effectively. This paper introduces a novel embedded system designed to autonomously label sensor data directly on IoT devices, thereby enhancing the efficiency of data collection methods. We present an integrated hardware and software solution equipped with specialized labeling sensors that streamline the capture and labeling of diverse types of sensor data. By implementing local processing with lightweight labeling methods, our system minimizes the need for extensive data transmission and reduces dependence on external resources. Experimental validation with collected data and a Convolutional Neural Network (CNN) model achieved a high classification accuracy of up to 91.67\%, as confirmed through 4-fold cross-validation. These results demonstrate the system’s robust capability to collect audio and vibration data with correct labels.

\keywords{Event Detection \and Time Series \and Sensor Data Collection \and Automated Labeling \and  Embedded Systems \and CNN \and Integrated Hardware System}
\end{abstract}

\section{Introduction}
\label{sec:introduction}
Event detection has become a popular topic in pervasive computing~\cite{yu2020spatiotemporal}, enabling intelligent systems to interpret environmental contexts and adapt configurations within various spaces, for example, offices or kitchens~\cite{vafeiadis2020audio,pandya2021ambient}. Traditional IoT methods often utilize multiple types of indirect sensor data, such as audio and vibrations~\cite{choudhary2022audio}, which are processed through Deep Learning (DL) models for event recognition.

Sufficiently labeled datasets are necessary to train DL models effectively~\cite{alzubaidi2023survey}. Typically, data streams are segmented and annotated with labels~\cite{bouchabou2021survey}. One common approach to collecting these datasets involves transmitting sensor data to the cloud~\cite{ghosh2020edge}, where labeling algorithms are applied~\cite{pivculjan2023machine}, or storing the data streams for subsequent manual labeling by human workers~\cite{fahy2022scarcity}. Both methods, however, introduce significant delays and dependencies on external resources.

Instead of transmitting data while collecting them, we propose a local processing approach. Given that IoT devices generally possess limited processing power~\cite{qian2022enhancing}, applying complex labeling algorithms in real-time during data collection poses significant challenges. To overcome this obstacle, we have developed a novel embedded system designed to collect and automatically label data using light-weight methods. This approach significantly reduces the need for continuous data transmission, aligning with the constraints of power, energy, and latency typical in the IoT context. The main contributions of this research include:

\begin{itemize}
    \item We designed an integrated hardware system equipped with various sensors and an SD card slot to facilitate on-device data and label storage. We also included additional labeling sensors to ensure accurate and efficient event detection.

    \item We developed software that features a predefined set of labels. The labeling process is automated through an interrupt- and threshold-based detection mechanism, significantly simplifying the computation required for label extraction.
    
    \item We validated the efficacy of our collected dataset through experiments with event classification using a  Convolutional Neural Network (CNN) model. On our custom dataset across three event types, our model achieved up to 91.67\% test accuracy, verified through 4-fold cross-validation.

\end{itemize}

The remainder of this paper is organized as follows: Section~\ref{sec:related_work} reviews relevant literature, setting the stage for our research. Section~\ref{sec:hardware_design} details our hardware design, while Section~\ref{sec:software_implementation} discusses the software implementation. Section~\ref{sec:results_evaluation} presents experiment setups and analyzes our findings. Finally, Section~\ref{sec:conclusion_future} concludes the paper and outlines directions for future research.


\section{Related Work}
\label{sec:related_work}

Previous studies predominantly relied on human involvement in the recording and labeling process, which not only complicates the procedure but also increases costs and the potential for errors during manual operations.

Specifically, Koch et al.~\cite{koch2022detection} manually controlled the start and stop of recordings for each event. While this method minimizes storage requirements, it introduces complexity and heightens the risk of human error. In contrast, Anand et al.~\cite{anand2022classification} implemented continuous data recording with post-collection labels based on timestamps and event types. This method simplifies the recording process but often accumulates large volumes of irrelevant data, leading to inefficient storage usage, especially when events are infrequent. Furthermore, while humans can feasibly label audio data by listening, this approach is impractical for vibration data.

In response to these challenges, our research introduces a novel automated system that significantly reduces the need for manual intervention by automating the collection of reference labels. Our approach utilizes additional sensors that only need light-weight computation to determine the event type locally. With an on-device approach, we are free from synchronization challenges and can efficiently capture the essential sensor data before and after an event occurs.


\section{Hardware System Design}
\label{sec:hardware_design}
\begin{figure}[!htb]
    \centering
    \includegraphics[width=.7\textwidth]{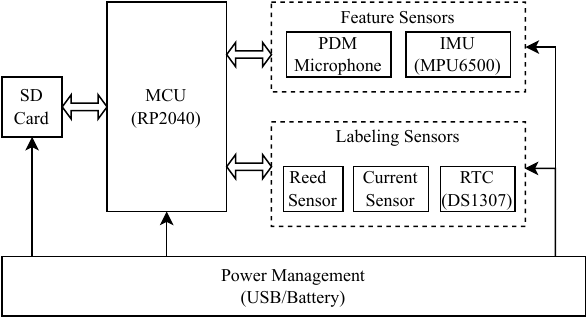}
    \caption{System Architecture of Elastic Node Sensor Logger}
    \label{fig:system_architecture}
\end{figure}

In this study, we propose a hardware platform named Elastic Node Sensor Logger. Figure \ref{fig:system_architecture} illustrates the architecture of our hardware platform, centered around the RP2040, an ARM Cortex-M0+ Microcontroller Unit (MCU) known for its low power consumption. In addition, its performance is sufficient for running FreeRTOS to make our multi-task scheduling easier than just using a bare-metal setup. This MCU also owns enough analog and digital I/O capabilities, which are crucial for managing the various sensors and storage modules incorporated into our system.

Our system includes two categories of sensors: feature and labeling sensors. 
The feature sensors are supposed to monitor the events indirectly. There is a Pulse Density Modulation (PDM) microphone for collecting audio data, and an Inertial Measurement Unit (IMU) programmed as an accelerator meter for collecting vibration data. Their sampling frequency is a parameter that the user can adjust. This configuration facilitates the generation of time series data, which is vital for training our DL models.

For event labeling, we utilize a reed sensor and a current sensor. The reed sensor detects door states by issuing a rising edge interrupt to the MCU when the door opens and a falling edge interrupt upon closing. Concurrently, the current sensor monitors the power consumption of a kettle. We use the analog-to-digital converter on MCU to detect the 'water has boiled' event based on a predefined current threshold (zero). In addition, a lower-power Real Time Clock (RTC) is embedded in the board to provide the timestamp for events. 

Additionally, an SD Card, connected to the MCU via the Serial Peripheral Interface, has been configured to operate at a writing speed with a clock frequency of up to 50 MHz. This high speed far exceeds the data acquisition rates from all sensors, ensuring that data logging remains efficient and does not hinder the system's overall performance.

The power management subsystem, including the MCP73833 for battery charging and the LM1117-3.3 for voltage regulation, ensures sufficient power utilization across all components. Given that the peak current consumption is estimated at 440 mW, a low-dropout regulator, LM1117-3.3, is sufficiently adequate for our power regulation needs, promoting system stability and efficiency.

\section{Software Implementation}
\label{sec:software_implementation}

\begin{figure}[!htb]
    \centering
    \includegraphics[width=1\textwidth]{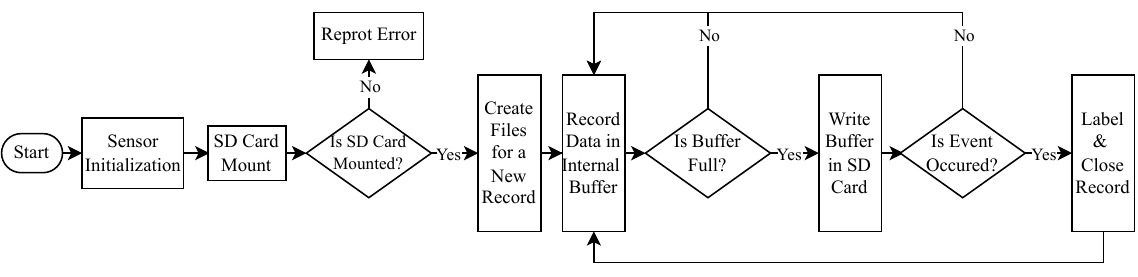}
    \caption{Main Loop of the Recording}
    \label{fig:main_loop}
\end{figure}

The main software loop, executed on the MCU, is depicted in Figure \ref{fig:main_loop}. At the outset, we initialize the sensor drivers and mount the SD card, setting the stage for data collection.

To optimize memory management and processing efficiency, we implement a ping-pong buffer strategy on the MCU for handling sensor data, a method akin to that described in \cite{microchip-ping}. Data is initially collected in the 'ping' buffer until it reaches capacity. At this point, data storage is switched to the 'pong' buffer. This cycle alternates to ensure continuous data acquisition. Upon filling either buffer, a Direct Memory Access (DMA) is triggered to transfer the data to the SD card, thereby offloading the data writing task from the MCU. This setup ensures that the SD card's write speed surpasses our data acquisition rate and provides ample buffer time to prevent overwriting and maintain data integrity.

The system has three event flags, two set by external interrupt callbacks and the third by a threshold-based trigger following ADC readings in a separate periodic task. This event detection logic is straightforward and computationally efficient, avoiding disruptions in data collection.

The system checks for flagged events once the DMA completes the data transfer from one buffer. If an event has been flagged, the corresponding label is immediately written to the SD card. The recording file may be closed promptly or left open for several seconds to capture additional post-event data, and the duration of the continuous recording is user-configurable.

Furthermore, our system utilizes an \textit{FatFs}\footnote{\url{https://github.com/elehobica/pico\_fatfs}} file system  to support up to four simultaneous file operations on the SD card. This capability allows for concurrent audio and vibration data recording, each stored in formats optimized for ease of access and analysis. Audio recordings are saved in WAV format for convenient review, while vibration data and event labels with timestamps are stored in separate CSV files, simplifying data management and enhancing accessibility.

\section{Experiments and Results}
\label{sec:results_evaluation}

\begin{figure}[!htb]
\centering
\begin{minipage}{0.49\textwidth}
    \centering
    \includegraphics[width=.9\linewidth]{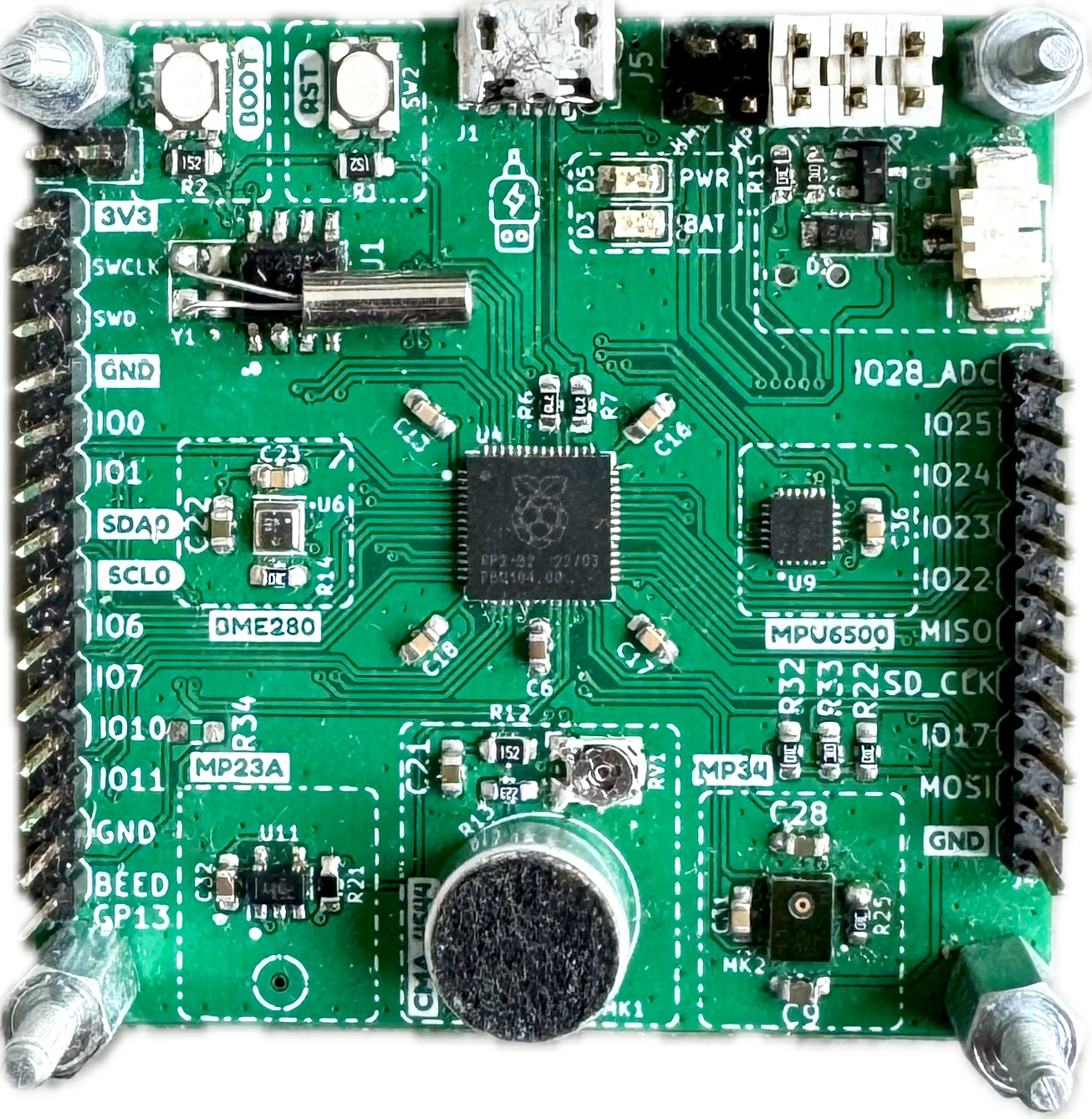}
    \caption*{(a) Front Side}
\end{minipage}
\hfill
\begin{minipage}{0.49\textwidth}
    \centering
    \includegraphics[width=.9\linewidth]{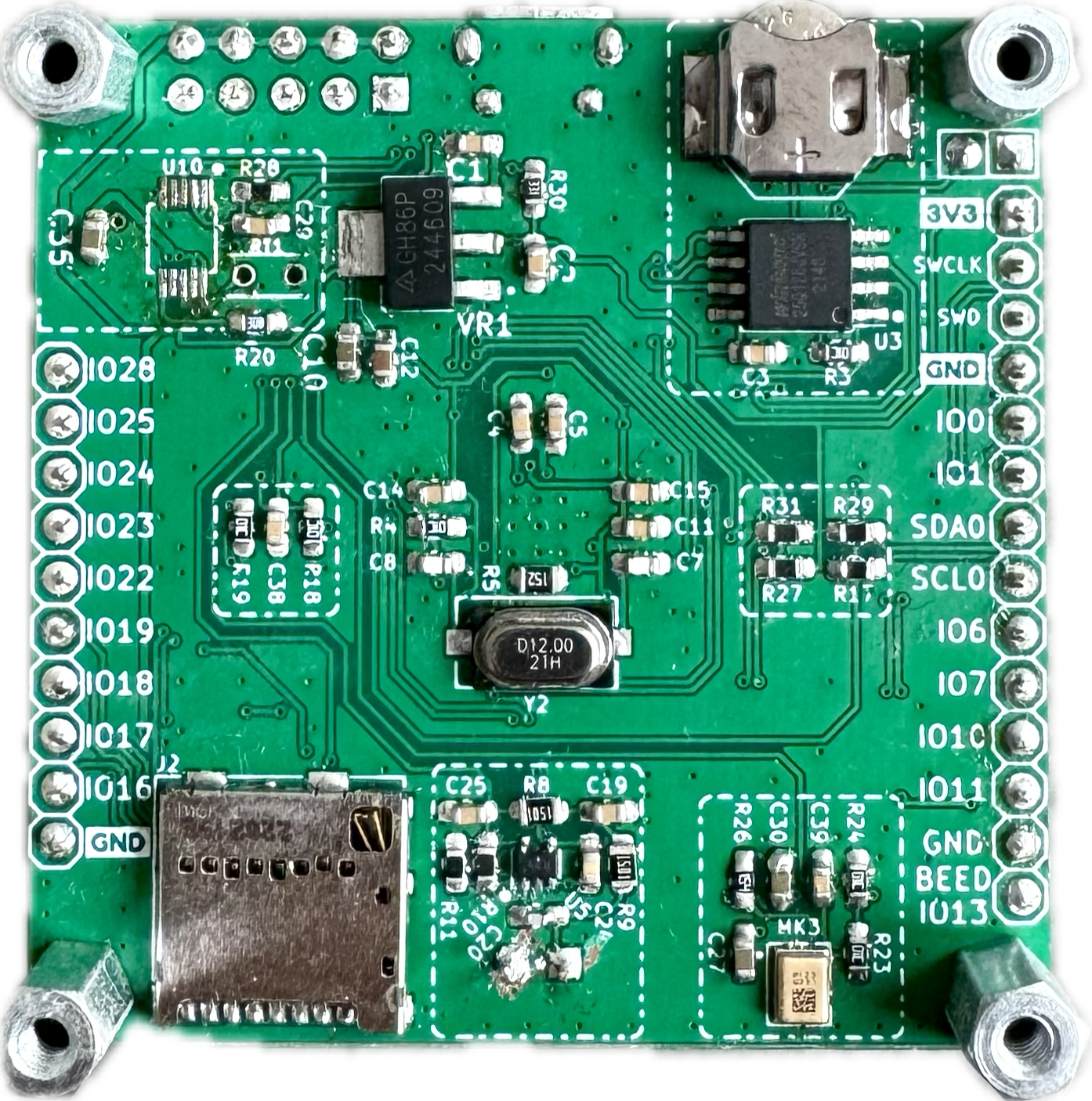}
    \caption*{(b) Back Side}
\end{minipage}
\caption{Elastic Node Sensor Logger}
\label{fig:elastic_node}
\end{figure}

Building upon the hardware system design outlined in Section~\ref{sec:hardware_design}, we successfully implemented the hardware platform as depicted in Figure~\ref{fig:elastic_node}. Utilizing this hardware and following the software implementation described in Section~\ref{sec:software_implementation}, we conducted multi-sensor data collection and labeling directly on our hardware platform. Subsequently, the collected dataset underwent preprocessing and validation on a desktop computer.

\subsection{Multi-Sensor Data Collection}

The data collection process solely requires the use of the Elastic Node Sensor Logger hardware. As mentioned in Section \ref{sec:software_implementation}, once the recording process initiates, our system simultaneously collects data from the microphone and the IMU sensor. The labeling sensors operate in the background as described. The audio data is captured at a sampling rate of 16 kHz in a mono-channel format. The vibration data, which includes three channels corresponding to acceleration, is collected at a sampling rate of 4 kHz. After starting the recording, the device operates autonomously for several hours. During this period, the user (operator) randomly engages in activities such as opening and closing doors and boiling water to generate event data. In total, we collected 106 samples from each type of sensor: 40 samples were associated with door opening events, 29 with door closing, and 37 with water boiling in a kettle.

\subsection{Data Preprocessing}

\begin{figure}[!htb]
\centering
\begin{minipage}{0.7\textwidth}
    \centering
    \includegraphics[width=\linewidth]{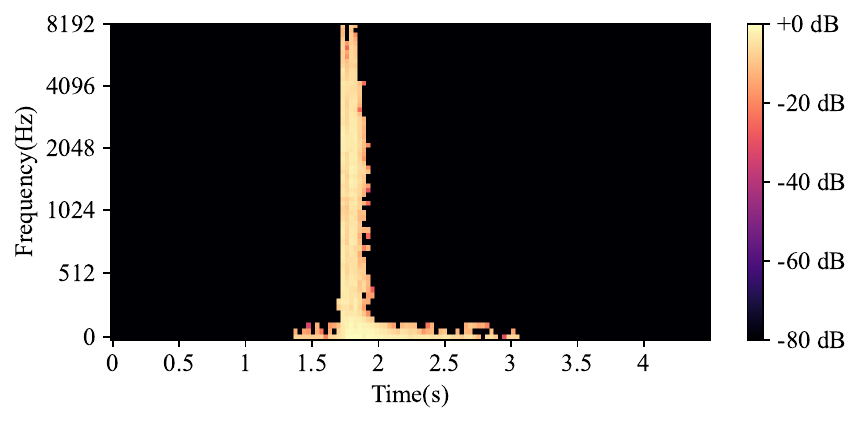}
    \caption*{(a) Close Door}
\end{minipage}
\vspace{10pt} 
\begin{minipage}{0.7\textwidth}
    \centering
    \includegraphics[width=\linewidth]{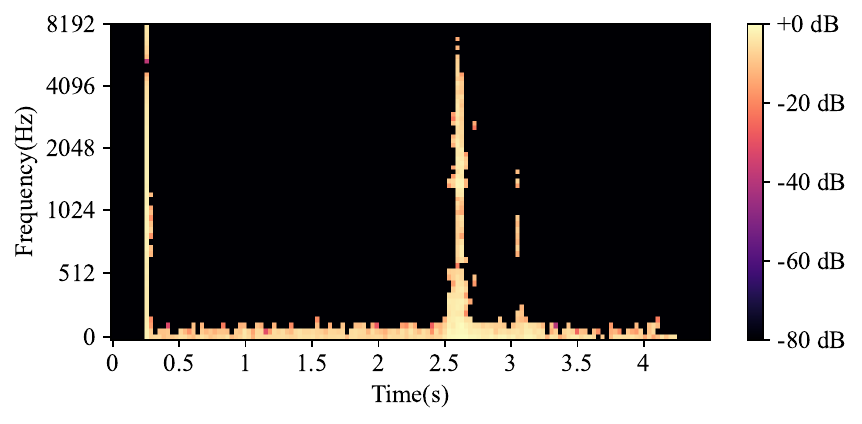}
    \caption*{(b) Open Door}
\end{minipage}
\vspace{10pt} 
\begin{minipage}{0.7\textwidth}
    \centering
    \includegraphics[width=\linewidth]{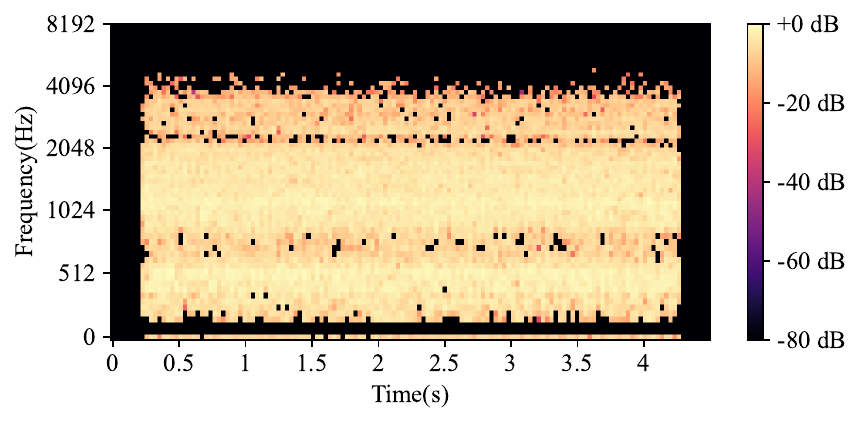}
    \caption*{(c) Water Boiling}
\end{minipage}
\caption{Mel Spectrograms of Different Event Labels}
\label{fig:audio_data_mel}
\end{figure}

\begin{figure}[!htb]
\centering
\begin{minipage}{0.7\textwidth}
    \centering
    \includegraphics[width=\linewidth]{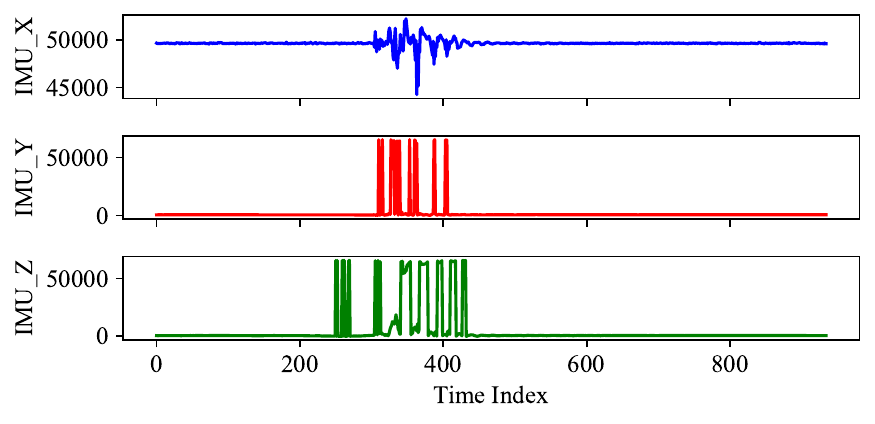}
    \caption*{(a) Close Door}
\end{minipage}
\begin{minipage}{0.7\textwidth}
    \centering
    \includegraphics[width=\linewidth]{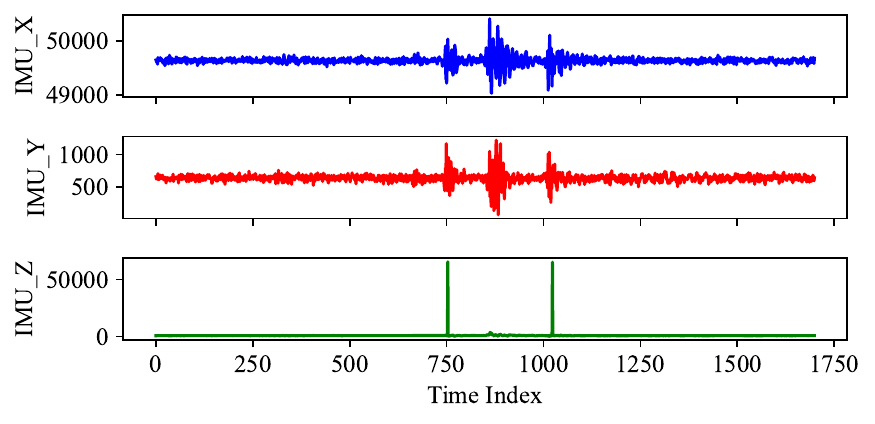}
    \caption*{(b) Open Door}
\end{minipage}
\begin{minipage}{0.7\textwidth}
    \centering
    \includegraphics[width=\textwidth]{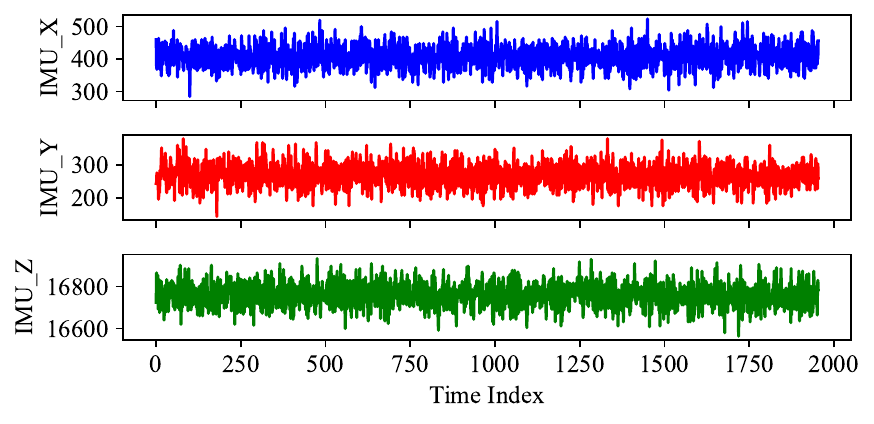}
    \caption*{(c) Water Boiling}
\end{minipage}
\caption{Visualization of vibration data with different event labels}
\label{fig:vibration_data}
\end{figure}

Before we fed our custom dataset to the model, audio and vibration data were preprocessed separately to accommodate their unique characteristics . Audio recordings were read from WAV files using torchaudio~\cite{yang2022torchaudio}. To standardize the lengths of these recordings, zero-padding was symmetrically applied to both ends to have the same length as the longest audio data in the dataset. For feature extraction, we transformed the recordings into Mel spectrograms using the following parameters: \(n_\text{mels} = 64\), \(n_\text{fft} = 1024\), with \(hop_\text{length}\) at default settings, and \(top_\text{db} = 80\) for dynamic range compression. Figure \ref{fig:audio_data_mel} displays these spectrograms for three events, showcasing their distinct spectral characteristics. For example, door-related events exhibit short-term peaks along the time axis and a broader range of frequency coverage compared to kettle-boiling events. Furthermore, distinguishable patterns are evident between door opening and closing events, such as the longer duration of closing events compared to opening events.

Vibration recordings, comprised of channel measurements, varied in length across samples. We addressed this by applying a zero-padding strategy similar to that used for audio data. In instances of missing values, we imputed these by calculating and using the mean of the respective dimension. Figure \ref{fig:vibration_data} displays the visualization of vibration data, categorized in terms of the three sample events. Distinct patterns are identifiable between the door-related events, and there is a clear difference between these and the water-boiling events in the time domain.

\subsection{Data Validation}

To verify the quality of the collected data and the accuracy of labeling, we conducted a three-class classification task based on audio and vibration data. We consider the quality of our collected dataset to be high if the collected data and labels can train a deep learning model to converge and achieve test high accuracy. We split the entire dataset into training, validation, and testing sets in a 3:1:1 ratio. Afterward, we utilized an oversampling strategy to balance the distribution of samples across different labels for both data types. Training and validation sets underwent a 4-fold cross-validation process. Moreover, we computed the mean and standard deviation based on the training set to normalize all datasets.

\begin{figure}[!htb]
\centering
\begin{tikzpicture}[node distance=0.9cm, auto, font=\scriptsize]
    \tikzset{
        block/.style={rectangle, draw, text centered, rounded corners, minimum height=1.5em},
        line/.style={draw, -Latex}
    }
    \node [block, text width=7em] (input) { Input Data};
    \node [block, text width=20em, below of=input, fill=blue!20] (conv1) {Conv Layer: 64 filters, Kernel Size: 3, Stride: 1}; 
    \node [block, text width=10em, below of=conv1, fill=gray!20] (act1) {ReLU + Batch Norm};
    \node [block, text width=20em, below of=act1, fill=blue!20] (conv2) {Conv Layer: 32 filters, Kernel Size: 3, Stride: 1};
    \node [block, text width=10em, below of=conv2, fill=gray!20] (act2) {ReLU + Batch Norm};
    \node [block, text width=12em, below of=act2, fill=pink!20] (pool) {Adaptive Average Pooling};
    \node [block, text width=12em, below of=pool, fill=yellow!20] (fc) {Fully Connected Layer};
    \node [block, text width=16em, below of=fc, fill=green!20] (output) {Output: Class Probabilities};

    \path [line] (input) -- (conv1);
    \path [line] (conv1) -- (act1);
    \path [line] (act1) -- (conv2);
    \path [line] (conv2) -- (act2);
    \path [line] (act2) -- (pool);
    \path [line] (pool) -- (fc);
    \path [line] (fc) -- (output);
\end{tikzpicture}
\caption{CNN Architecture for Time-series Data Classification}
\label{fig:cnn_architecture}
\end{figure}
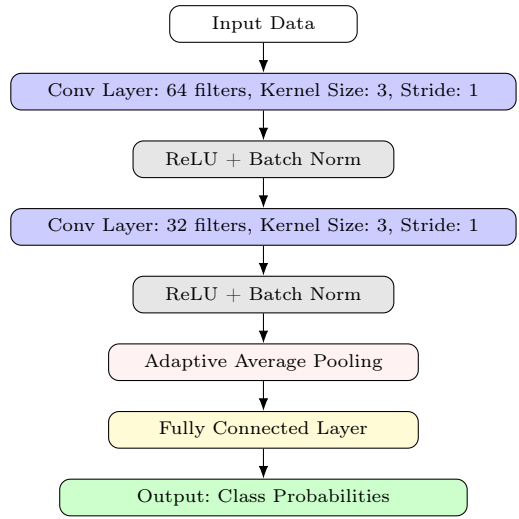

We adopted a simple CNN model for event classification under the PyTorch framework. As depicted in Figure \ref{fig:cnn_architecture}, this model features two convolutional layers, with the first layer having 64 filters and the second 32 filters using a kernel size of 3 and stride of 1. Each convolutional layer is equipped with Rectified Linear Unit (ReLU) activation function and batch normalization, enhancing the model's learning efficiency. They are then followed by an adaptive average pooling layer that reduces dimensionality, preparing the output for the final classification stage. The processed data is fed into a fully connected layer, classifying events. 

We configured our model training using the \emph{Adam} optimizer, setting hyperparameters to \( \beta_1 = 0.9 \), \( \beta_2 = 0.98 \), and \( \epsilon = 10^{-9} \). The training initiated with a learning rate of 0.001, which we dynamically adjusted using a scheduler that modified the rate at a step size of 3 with a decay factor \( \gamma \) of 0.5. We opted for cross-entropy error as the loss function to train and evaluate the model's performance. To enhance the robustness of our training process, we conducted 100 experiments, each comprising 50 epochs, and incorporated an early stopping mechanism to mitigate the risk of overfitting. We used accuracy as the primary evaluation metric complemented by a confusion matrix to provide detailed insights into the model's performance across different events.

\begin{figure}[!htb]
    \centering
    \includegraphics[width=.7\textwidth]{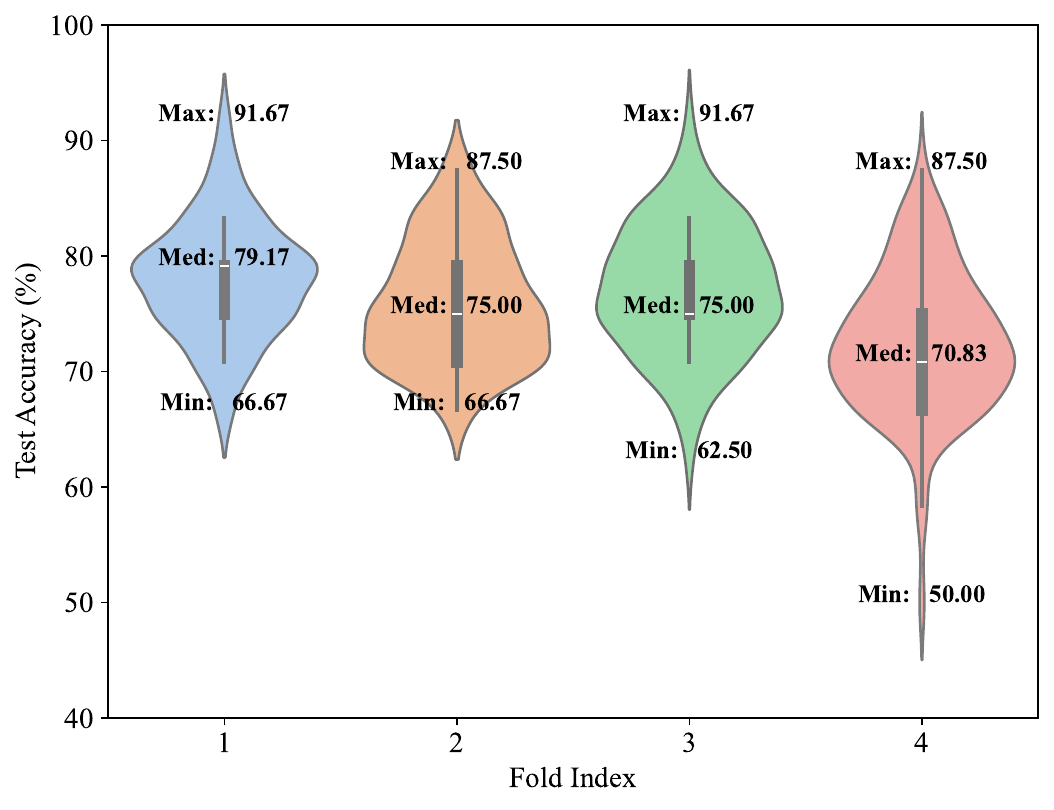}
    \caption{Audio Data: Test Accuracy (\%) across Different Folds}
    \label{fig:audio_data_results}
\end{figure}
\begin{figure}[!htb]
    \centering
    \includegraphics[width=.7\textwidth]{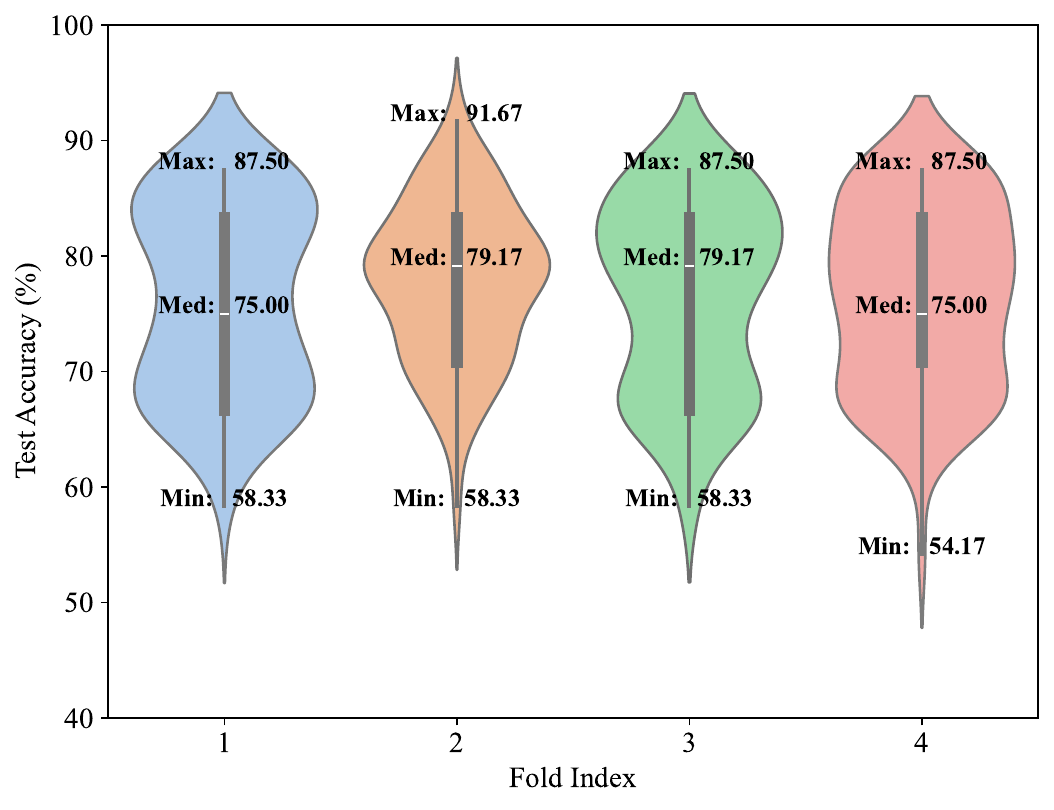}
    \caption{Vibration Data: Test Accuracy (\%) across Different Folds}
    \label{fig:vib_data_results}
\end{figure}

Figure \ref{fig:audio_data_results} illustrates the distribution of test accuracy for audio data across different validation folds. The observed minimum accuracy ranged from 50.00\% to 66.67\%. Despite these variations, the model demonstrates strong potential, achieving maximum accuracy up to 91.67\% in folds 1 and 3, and 87.50\% in folds 2 and 4. The median accuracy, spanning from 70.83\% to 79.17\%, suggests that the model generally maintains high-performance levels. Figure \ref{fig:vib_data_results} presents the test accuracy for vibration data, which also exhibits variability with minimum accuracy between 54.17\% and 58.33\%. The model reaches a high accuracy of up to 91.67\% in fold 2 and consistently above 87.50\% in the other folds. The median accuracy, consistently between 75.00\% and 79.17\%, indicates a reliable performance. 
\begin{figure}[!htb]
\centering
\begin{minipage}{0.485\textwidth}
    \centering
    \includegraphics[width=\linewidth]{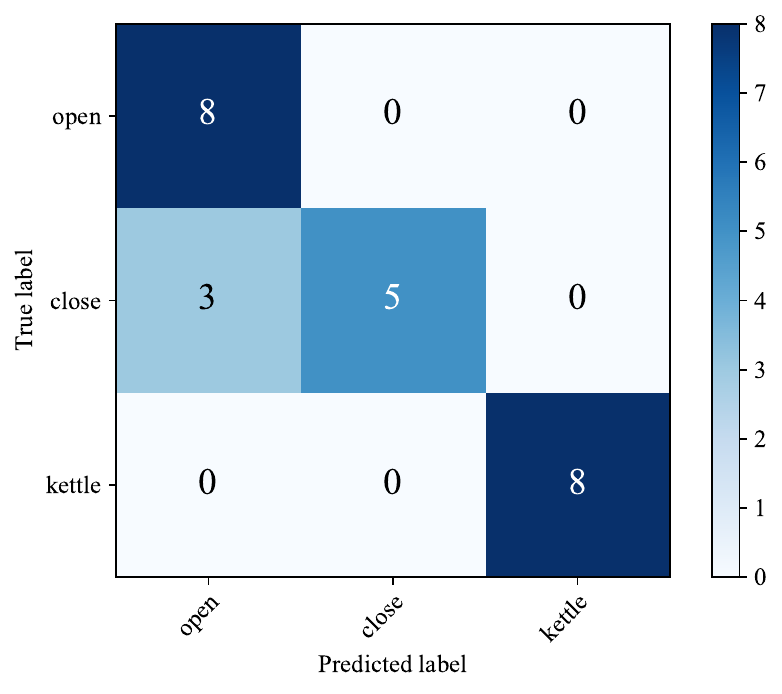}
    \caption*{(a) Audio Data}
\end{minipage}
\hfill
\begin{minipage}{0.485\textwidth}
    \centering
    \includegraphics[width=\linewidth]{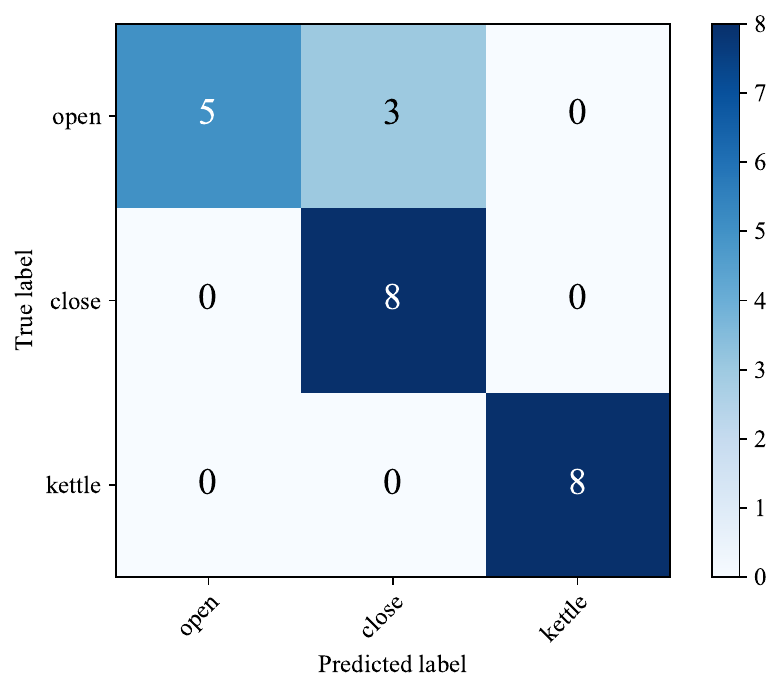}
    \caption*{(b) Vibration Data}
\end{minipage}
\caption{Confusion Matrix With Test Accuracy 87.5\%}
\label{fig:confusion_matrix}
\end{figure}

To understand the limitations of our system and identify potential areas for improvement, we conducted a detailed analysis of a trained model with a test accuracy of 87.5\% using a confusion matrix. Figure \ref{fig:confusion_matrix} (a) displays the confusion matrix for the model trained with audio data. It reveals that all samples of water boiling and door opening are correctly classified, although three samples of door closing were misclassified as door opening. Similarly, Figure \ref{fig:confusion_matrix} (b), which pertains to the model trained with vibration data, shows that only three samples of door-opening events were misclassified as door-closing events.

It is important to note that although we collected audio and vibration data concurrently, the models were trained separately on each data type. Integrating both audio and vibration data as inputs for the models could potentially enhance accuracy further, particularly in applications requiring high precision. However, investigating this integrated approach is beyond the scope of this paper.

In summary, the quality of the collected data and labels has proven sufficient for CNN to learn and differentiate between event classes effectively. While the classification accuracy from vibration data is slightly lower than that from audio data, this outcome was anticipated due to the inherent challenges associated with vibration signal classification. Notably, the consistency of our sensor data has been validated across four-folds, confirming the effectiveness of our system in capturing classifiable features across three distinct event types.

\section{Conclusion and Future Work}
\label{sec:conclusion_future}
Our study successfully developed a robust approach for autonomously labeling sensor data directly on IoT devices. Experiments demonstrated that our models achieved up to 91.67\% test accuracy in controlled settings, highlighting the high quality of our sensor data and the reliability of our labeling approach. This method significantly improves the feasibility of collecting and processing large-scale IoT data in diverse field environments, enhancing efficiency and accuracy.

In our ongoing efforts to enhance event detection capabilities, we plan to integrate additional types of feature sensors into our system. This expansion will enable the support and recognition of a broader array of event types, further improving the versatility and applicability of our solution in diverse scenarios. By broadening the sensor array, we aim to capture more comprehensive feature data from the device surroundings, significantly refining our system's responsiveness and accuracy in real-world applications.

\section*{Acknowledgments} 
The authors gratefully acknowledge the financial support provided by the Federal Ministry for Economic Affairs and Climate Action of Germany for the RIWWER project (01MD22007C).

\bibliographystyle{splncs04_unsort}
\bibliography{reference}

\begin{thebibliography}{10}
\providecommand{\url}[1]{\texttt{#1}}
\providecommand{\urlprefix}{URL }
\providecommand{\doi}[1]{https://doi.org/#1}

\bibitem{yu2020spatiotemporal}
Yu, M., Bambacus, M., Cervone, G., Clarke, K., Duffy, D., Huang, Q., Li, J., Li, W., Li, Z., Liu, Q., et~al.: Spatiotemporal event detection: a review. International Journal of Digital Earth  \textbf{13}(12),  1339--1365 (2020)

\bibitem{vafeiadis2020audio}
Vafeiadis, A., Votis, K., Giakoumis, D., Tzovaras, D., Chen, L., Hamzaoui, R.: Audio content analysis for unobtrusive event detection in smart homes. Engineering Applications of Artificial Intelligence  \textbf{89},  103226 (2020)

\bibitem{pandya2021ambient}
Pandya, S., Ghayvat, H.: Ambient acoustic event assistive framework for identification, detection, and recognition of unknown acoustic events of a residence. Advanced Engineering Informatics  \textbf{47},  101238 (2021)

\bibitem{choudhary2022audio}
Choudhary, P., Kumari, P., Goel, N., Saini, M.: An audio-seismic fusion framework for human activity recognition in an outdoor environment. IEEE Sensors Journal  \textbf{22}(23),  22817--22827 (2022)

\bibitem{alzubaidi2023survey}
Alzubaidi, L., Bai, J., Al-Sabaawi, A., Santamar{\'\i}a, J., Albahri, A.S., Al-dabbagh, B.S.N., Fadhel, M.A., Manoufali, M., Zhang, J., Al-Timemy, A.H., et~al.: A survey on deep learning tools dealing with data scarcity: definitions, challenges, solutions, tips, and applications. Journal of Big Data  \textbf{10}(1), ~46 (2023)

\bibitem{bouchabou2021survey}
Bouchabou, D., Nguyen, S.M., Lohr, C., LeDuc, B., Kanellos, I.: A survey of human activity recognition in smart homes based on {IoT} sensors algorithms: taxonomies, challenges, and opportunities with deep learning. Sensors  \textbf{21}(18), ~6037 (2021)

\bibitem{ghosh2020edge}
Ghosh, A.M., Grolinger, K.: Edge-cloud computing for internet of things data analytics: embedding intelligence in the edge with deep learning. IEEE Transactions on Industrial Informatics  \textbf{17}(3),  2191--2200 (2020)

\bibitem{pivculjan2023machine}
Pi{\v{c}}uljan, N., Car, {\v{Z}}.: Machine learning-based label quality assurance for object detection projects in requirements engineering. Applied Sciences  \textbf{13}(10), ~6234 (2023)

\bibitem{fahy2022scarcity}
Fahy, C., Yang, S., Gongora, M.: Scarcity of labels in non-stationary data streams: a survey. ACM Computing Surveys (CSUR)  \textbf{55}(2),  1--39 (2022)

\bibitem{qian2022enhancing}
Qian, C., Ling, T., Schiele, G.: Enhancing energy-efficiency by solving the throughput bottleneck of {LSTM} cells for embedded {FPGA}s. In: Joint European Conference on Machine Learning and Knowledge Discovery in Databases. pp. 594--605. Springer (2022)

\bibitem{koch2022detection}
Koch, M., Schlenke, F., Kohlmorgen, F., Kuller, M., Bauer, J., Woehrle, H.: Detection and classification of human activities using distributed sensing of environmental vibrations. In: 2022 IEEE International Conference on Omni-layer Intelligent Systems (COINS). pp.~1--6. IEEE (2022)

\bibitem{anand2022classification}
Anand, J., Koch, M., Schlenke, F., Kohlmorgen, F., W{\"o}hrle, H.: Classification of human indoor activities with resource constrained network architectures on audio data. In: 2022 IEEE 5th International Conference and Workshop {\'O}buda on Electrical and Power Engineering (CANDO-EPE). pp. 000157--000162. IEEE (2022)

\bibitem{microchip-ping}
Prasad, M.: Ping-pong buffers. \url{https://onlinedocs.microchip.com/pr/GUID-324A966D-1464-4B35-A7D1-DCAE052AC22C-en-US-3/index.html?GUID-B6995A5F-E06B-4071-893E-BBC60082F576} (2024)

\bibitem{yang2022torchaudio}
Yang, Y.Y., Hira, M., Ni, Z., Astafurov, A., Chen, C., Puhrsch, C., Pollack, D., Genzel, D., Greenberg, D., Yang, E.Z., et~al.: Torchaudio: building blocks for audio and speech processing. In: ICASSP 2022-2022 IEEE International Conference on Acoustics, Speech and Signal Processing (ICASSP). pp. 6982--6986. IEEE (2022)

\end{thebibliography}
\end{document}